\documentclass{article}

\usepackage{arxiv}

\usepackage[utf8]{inputenc} 
\usepackage[T1]{fontenc}    
\usepackage{hyperref}       
\usepackage{url}            
\usepackage{booktabs}       
\usepackage{amsfonts}       
\usepackage{nicefrac}       
\usepackage{microtype}      
\usepackage{cleveref}       
\usepackage{lipsum}         
\usepackage{graphicx}
\usepackage{doi}
\usepackage{graphicx}
\usepackage{booktabs}
\usepackage{fontenc}
\usepackage{makecell}
\usepackage{bm}
\usepackage{float}

\title{GNNFormer: A Graph-based Framework for Cytopathology Report Generation}

\author{ \\
    \textbf{Yang-Fan Zhou}, \textbf{Kai-Lang Yao} and \textbf{Wu-Jun Li}\\
	National Key Laboratory for Novel Software Technology\\
	Department of Computer Science and Technology, Nanjing University, China\\	\texttt{\{zhouyangfan,yaokl\}@smail.nju.edu.cn,  liwujun@nju.edu.cn} \\
}

\renewcommand{\undertitle}{Technical Report}

\hypersetup{
pdftitle={GNNFormer: A Graph-based Framework for Cytopathology Report Generation},
pdfsubject={Artificial Intelligence},
pdfauthor={yangfan zhou},
pdfkeywords={Cytopathology, Image Caption Generation},
}

\begin{document}
\maketitle

\begin{abstract}
	Cytopathology report generation is a necessary step for the standardized examination of pathology images. However, manually writing detailed reports brings heavy workloads for pathologists. To improve efficiency, some existing works have studied automatic generation of cytopathology reports, mainly by applying image caption generation frameworks with visual encoders originally proposed for natural images. A common weakness of these works is that they do not explicitly model the structural information among cells, which is a key feature of pathology images and provides significant information for making diagnoses. In this paper, we propose a novel graph-based framework called \mbox{GNNFormer}, which seamlessly integrates graph neural network~(GNN) and Transformer into the same framework, for cytopathology report generation. To the best of our knowledge, \mbox{GNNFormer} is the first report generation method that explicitly models the structural information among cells in pathology images. It also effectively fuses structural information among cells, fine-grained morphology features of cells and background features to generate high-quality reports.
   Experimental results on the NMI-WSI dataset show that \mbox{GNNFormer} can outperform other state-of-the-art baselines.
\end{abstract}


\section{Introduction}
\label{sec:intro}

Pathology image is one of the major references for the diagnosis of tumors. By carefully examining the huge-sized pathology images, pathologists give conclusions of tumors' existence and status. During the examination, pathologists also need to pick out several representative image patches and write cytopathology reports for them. These reports should describe in details whether the cells in these patches are cancerous. This is a significant part of the examination, since these patches and corresponding reports provide strong evidence for the final diagnostic conclusions. However, picking out typical patches from huge-sized pathology images and writing detailed cytopathology reports could be very time-consuming. Hence, developing a model that automatically diagnoses pathology image patches and generates cytopathology reports could hopefully increase the efficiency of pathology image examination.

Although many methods have been proposed for the classification or segmentation of cells~\cite{graham2019hover,wang2020single} and lesions~\cite{liang2018weakly,lu2021ai} in pathology images, only a few study the generation of cytopathology reports. 
They mainly focus on improving the report generator according to the characteristics of medical reports, while applying a visual encoder that is originally proposed for natural images to pathology images. 
Authors in ~\cite{conf/cvpr/ZhangXXMY17} employs ResNet~\cite{he2016deep} as visual encoder, and uses multiple long short term memory~(LSTM) networks to describe different symptoms in parallel. Authors in~\cite{jing2018automatic} propose a multi-task medical report generalization framework that jointly performs the prediction of tags and the generation of paragraphs. 
\begin{figure}
    \centering
	\includegraphics[width=0.8\linewidth]{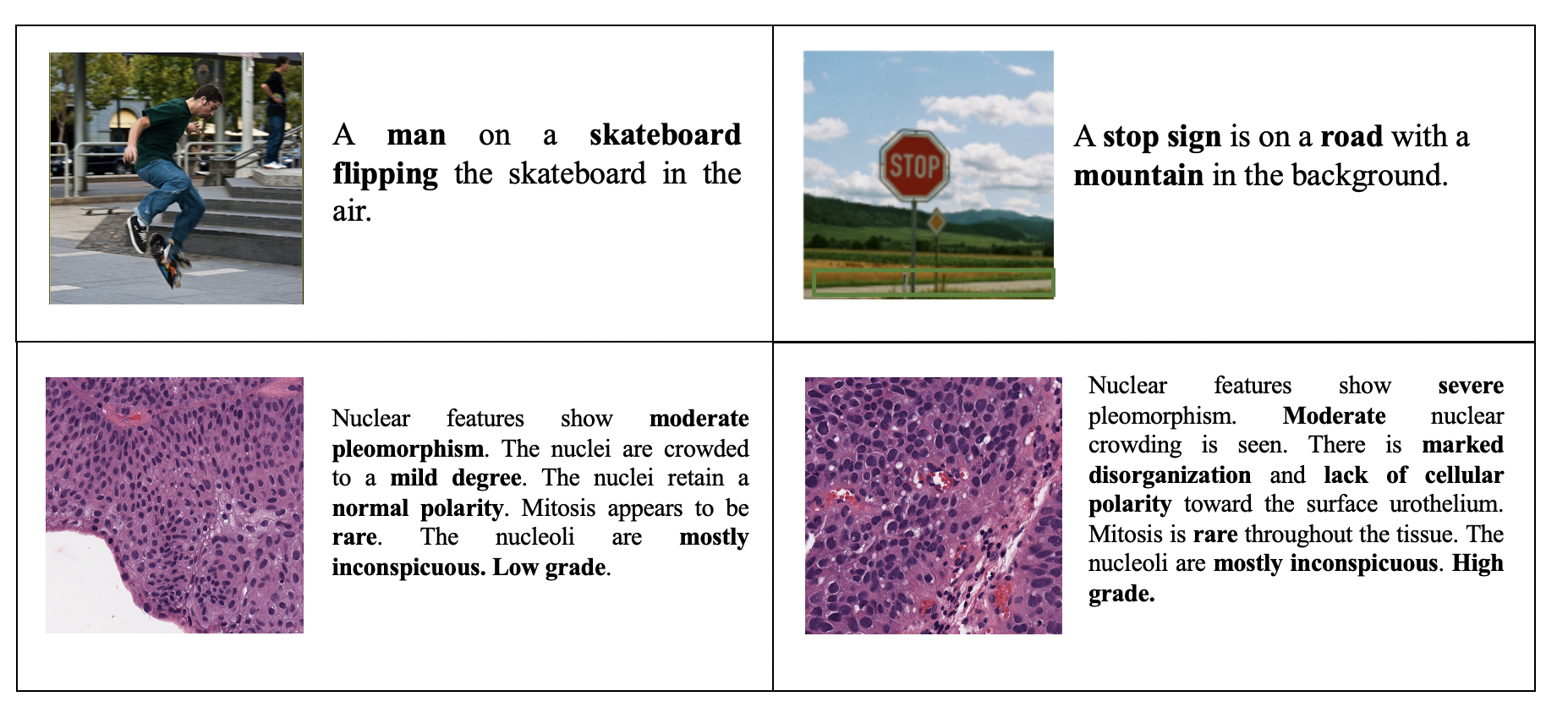}
	\caption{Examples of natural images~(up) and pathology image patches~(down), with the corresponding caption or report. For natural images, the caption describes several main objects and their status. While for pathology image patches, the report focuses on the joint status of numerous cells and gives summary conclusions.} \label{diff}
	\vskip -0.1in
\end{figure}
A common weakness of these methods is that they do not explicitly model the structural information among cells, which is a key feature of pathology images.
Unlike natural images, pathology images mainly contain cell clusters that are composed of numerous nuclei with similar features. Figure~\ref{diff} gives examples of natural images and pathology images with the corresponding captions and reports. We can find that information in cytopathology reports is mainly concluded from critical visual morphology as well as structural information among cells. Particularly, structural information among cells is essential for drawing diagnosis conclusions. For diagnosing the urothelial carcinoma shown in Figure~\ref{diff}, distributions of nuclear pleomorphism state, cell crowding, cell polarity, mitosis, and nucleoli prominence are primary references needed to be concerned~\cite{zhang2019pathologist}. In addition, disruption in the cohesion of architecture between nuclei and other primitives of the same family shows a high possibility of presence for a number of different cancers~\cite{lu2018feature}. Hence, explicitly modeling the structural information among cells is important for generating high-quality cytopathology reports.

In this paper, we propose a novel graph-based framework called GNNFormer, which seamlessly integrates graph neural network~(GNN)~\cite{scarselli2008graph} and Transformer~\cite{vaswani2017attention} into the same framework, for cytopathology report generation. 
The main contributions of this paper are outlined as follows: 
\begin{itemize}
    \item To the best of our knowledge, GNNFormer is the first report generation method that explicitly models structural information among cells in pathology images.
    \item GNNFormer effectively fuses structural information among cells, fine-grained morphology features of cells and background features of images to generate high-quality cytopathology reports.
     \item GNNFormer can generate visualization of cells' importance scores for report generation, providing intuitive interpretation for pathologists.
    \item Experiments on the NMI-WSI dataset~\cite{zhang2019pathologist} show that GNNFormer can outperform other state-of-the-art baselines. 
   
\end{itemize}

\begin{figure*}[t]
    \centering
	\includegraphics[width=0.9\linewidth]{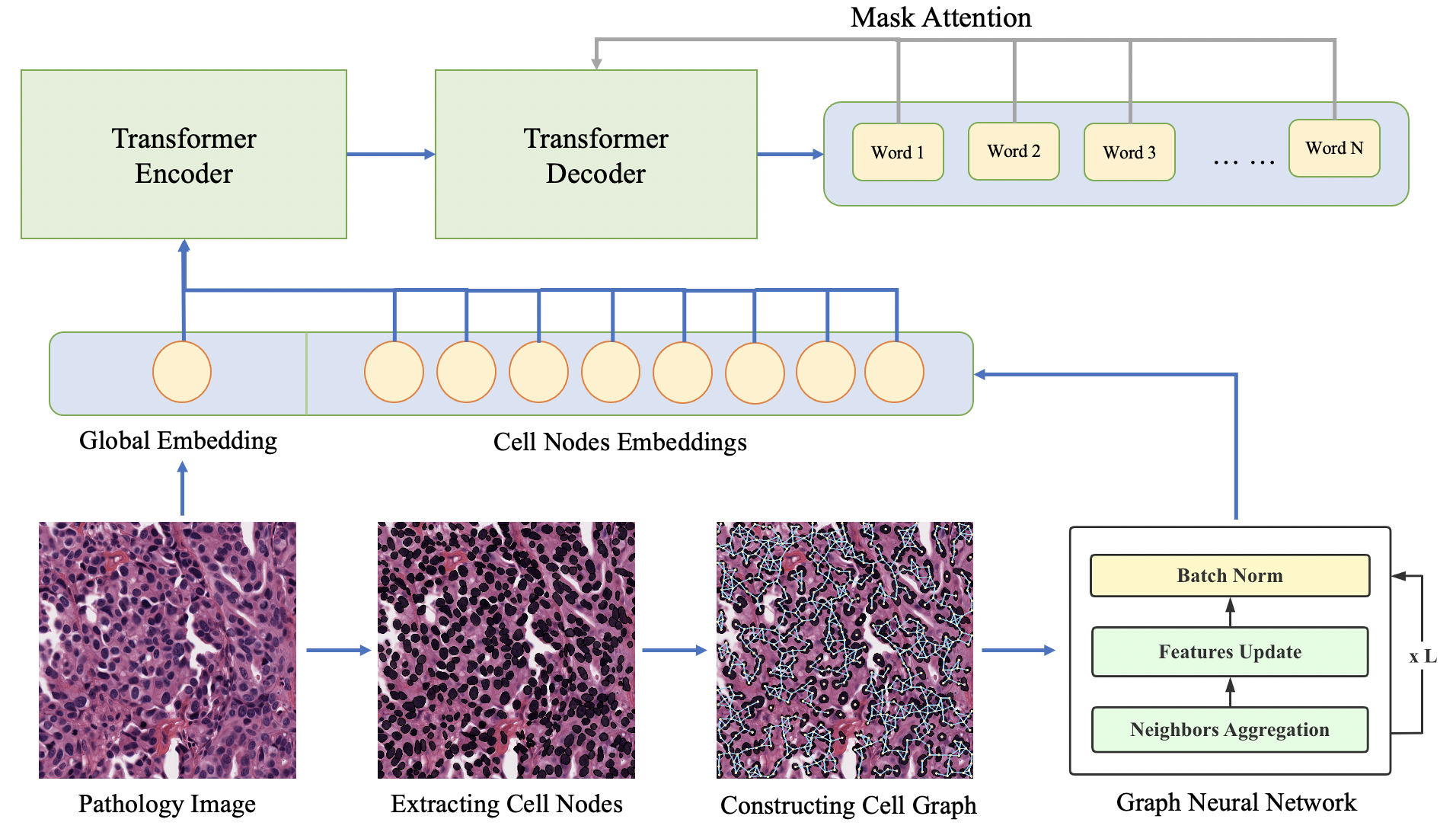}
	\caption{The framework of GNNFormer. GNNFormer first extracts cell nodes from the input pathology image and constructs a cell graph. Then it adapts a GNN module to capture structural information among cells. At last, an encoder-decoder Transformer module is used to generate a cytopathology report based on visual embeddings. } \label{framework}
\end{figure*}

\section{Related Works}
\label{sec:related}
\subsection{Image Caption Generation}
Image caption generation aims to generate natural language descriptions for input images. The most widely adopted model structures for image caption generation typically contain a visual encoder to extract information from the input images and a language model to generate captions. 
Most works adopt convolution neural networks~(CNNs) as visual encoders. For example, early works~\cite{conf/cvpr/VinyalsTBE15,Donahue_2015_CVPR} try to extract visual feature vectors from top layers of CNNs. The work in~\cite{xu2015show} proposes to use an attention layer to control the area of focus before generating each word. In~\cite{anderson2018bottom}, Faster-RCNN is used to detect salient regions in the images. 
Several works also build graphs on salient regions\cite{conf/eccv/YaoPLM18, conf/cvpr/YangTZC19}, and they are mostly concerned about building graphs of major objects that accurately capture important semantic and spatial relationships. These methods cannot be directly applied to pathology images, since the number of cells in pathology images are too large and the diagnostic conclusions should be made according to the structural information among numerous cells in clusters rather than relationship between single cells.
As for the language model, recurrent neural network~(RNN) has been widely used in caption generation~\cite{conf/cvpr/VinyalsTBE15, Donahue_2015_CVPR, xu2015show, anderson2018bottom}. Lately, Transformer~\cite{conf/acl/SoricutDSG18,conf/nips/HerdadeKBS19,cornia2020meshed} has  also been widely adopted due to its good performance on natural language processing tasks. 

\subsection{Computational Pathology}
Computational pathology~(CP) is a research field that uses computer algorithms to automatically analyze the patterns of pathology images. In recent years, CP has been developing rapidly due to the superior feature-capturing ability of deep learning models and the significant improvement of microscopic scanning devices. Works in~\cite{wang2016deep,garcia2017automatic,li2018deep,lu2021ai} apply CNN to extract image features and perform classification of pathology image patches, including cell types, lesion existence and tumor grading. Works in~\cite{raju2020graph,anand2020histographs} build cell graphs to improve the classification performance, showing cell graphs' potential in capturing essential structural information. 
Works in~\cite{liang2018weakly,graham2019mild,zhang2022deep} study the segmentation of lesion areas, including tumor and inflammation, with semantic segmentation models. 
For the segmentation and classification of cells, the work in~\cite{graham2019hover} adapts three segmentation branches to respectively predict cell existence, distance to cell border and class of each pixel. The work in~\cite{wang2020single} unifies instance segmentation and classification into a multi-task learning framework. Besides giving detailed diagnoses on patches, works in~\cite{zhu2017wsisa,mobadersany2018predicting,lu2021data} ensemble patch features to form global feature of huge-sized Whole-Slide-Image~(WSI). Works in~\cite{li2018graph,chen2021whole} build graphs of patches to extract the hierarchical information.

Several works also try to generate diagnostic reports with natural language, for providing more direct and intuitive reference to pathologists. They mainly focus on improving the report generator according to the characteristics of medical reports, while applying a visual encoder that is originally proposed for natural images to pathology images. MDNet~\cite{conf/cvpr/ZhangXXMY17} employs ResNet~\cite{he2016deep} as visual encoder, and uses multiple LSTM networks to describe different symptoms in parallel. It also introduces an attention module and an auxiliary attention sharpening module to highlight key regions before the LSTM module. In~\cite{jing2018automatic}, a multi-task medical report generation framework is proposed to jointly perform the prediction of tags and the generation of paragraphs. These methods do not explicitly model the structural information among cells, which may result in the loss of significant information for making diagnosis. This motivates the work in this paper.

\section{Method}
We use boldface uppercase letters, such as $\bm{A}$, to denote matrices. Boldface lowercase letters, such as $\bm{a}$, are used to denote column vectors. Scalars are denoted by ordinary lowercase letters, such as $a$. Sequences are denoted by uppercase letters, such as $A$. Squiggly letters such as $\mathcal{A}$ are used to denote sets. 
We use $\bm{I}\in \mathbb{R}^{h\times w}$ to denote an input pathology image, where $h$ and $w$ are the height and width of the image, respectively.

The framework of GNNFormer is illustrated in Figure~\ref{framework}. \mbox{GNNFormer} first constructs a cell graph for each pathology image to explicitly model the structural information among cells. Cell nodes embeddings are initially generated by extracting cells' morphology features. Based on the constructed cell graph, GNNFormer uses a graph neural network~(GNN) to fuse structural information among cells into cell nodes embeddings. With cell nodes embeddings and global embedding of the image as input, GNNFormer adopts an encoder-decoder Transformer to generate reports in natural language. The following content will introduce the details about the three steps of the whole process of \mbox{GNNFormer}: cell graph construction, embedding generation and report generation.

\subsection{Cell Graph Construction}\label{sec:cgc}
We first construct a cell graph for each pathology image to explicitly model the structural information among cells, before the training of GNNFormer. In particular, we construct the cell graphs by following the framework in~\cite{jaume2021quantifying}. Firstly, we apply the HoverNet~\cite{graham2019hover} pre-trained on the PanNuke dataset~\cite{gamper2019pannuke} to segment and extract cell nodes from image $\bm{I}$. No further finetuning step is required to achieve good segmentation results in our experiments. Applying this cell segmentation method can greatly reduce the requirement for cell-level annotation of new datasets. Secondly, we extract visual and spatial features for cell nodes. We crop a patch of size $36\times 36$ pixels around each cell node, and generate the patch embedding using a ResNet34~\cite{he2016deep} pre-trained on ImageNet~\cite{deng2009imagenet}. Patch embeddings are treated as embedding for cell nodes. Note that embeddings generated in this step are used for graph construction only, since the involved model parameters are not trained with the report generation task. In order to introduce the spatial information, we also concatenate the central coordinates of each cell node to the corresponding embedding. Finally, we construct a cell graph $\mathcal{G}=\{\mathcal{V},\mathcal{E}\}$ for each pathology image by applying the $k$-nearest neighbor~(kNN) algorithm on the embeddings of cell nodes. $\mathcal{V}$ denotes the set of cell nodes and $\mathcal{E}$ denotes the set of edges. In the resulting cell graphs, cell nodes with similar morphology and adjacent positions are connected by edges.

\subsection{Embedding Generation}
After constructing the cell graphs, we generate the embeddings of cell nodes and input images for the report generation task. For embedding generation of cell nodes, we first use a trainable CNN to process the extracted cell nodes in Section~\ref{sec:cgc} and obtain the output $\bm{H}^{(0)}=[\bm{h}_1^{(0)}, \bm{h}_2^{(0)},\,\cdots \,,\bm{h}_{|\mathcal{V}|}^{(0)}]^\top$, which captures morphology features of cell nodes. In this way, the network can learn to extract essential morphology features for report generation during training. Then, we adopt a GNN model, namely graph isomorphism network~(GIN)~\cite{DBLP:conf/iclr/XuHLJ19}, to extract the structural information of cell graphs. Specifically, one layer of the GNN model is defined as follows:
\begin{equation}	\bm{h}_i^{(\ell)} =\mathrm{MLP}^\ell\left((1+\epsilon^{(\ell)})\bm{h}_i^{(\ell-1)}+\frac{1}{|\mathcal{N}_i|}\sum_{j\in \mathcal{N}_i}{\bm{h}_j^{(\ell-1)}}\right),
\end{equation}
where $\bm{h}_i^{(\ell)}$ denotes the hidden embedding of cell node $i$ in a cell graph at the $\ell$th layer. $\epsilon^{(\ell)}$ is a learnable parameter to distinguish central nodes from neighbors. $\mathcal{N}_i$ denotes the set of neighbors of cell node $i$. $\mathrm{MLP}^{\ell}$ denotes a multi-layer perceptron. $\bm{H}^{(0)}$ is the input of the GNN model. We collect hidden cell nodes embeddings from all GNN layers to generate the final cell nodes embeddings. The embeddings will fuse the morphology features of cells with structural information among cells. The process is defined as follows:
\begin{equation}\label{eq:cn-emb}
	\bm{h}_i^{o} = \sum_{\ell}\alpha_\ell \bm{h}_i^{(\ell)},
\end{equation}
where $\bm{h}_i^{o}$ denotes the final embedding of cell node $i$, $\alpha_\ell$ is the weight of hidden embeddings at the $\ell$th layer. We set $\bm{\alpha}=\{\alpha_\ell\}$ as a learnable parameter to allow the model to learn the best weights. 

For embedding generation of the input images, we use a CNN to obtain a global embedding $\bm{e}$ for the input image $\bm{I}$, where $\bm{e}$ summarizes the information of image background in $\bm{I}$. After that, we concatenate the global embedding $\bm{e}$ with the sequence of the extracted cell nodes embeddings. In this way, information including background colors, texture and global structures can be captured by the generated embedding, which will be used as input for the following report generation step.

\subsection{Report Generation}
After the embeddings are generated, we use a Transformer structure for report generation due to its superior capability in global information extraction. The overall process of report generation mainly consists of two stages. In the first stage, we use an encoder Transformer to integrate information provided by the global embedding $\bm{e}$ of the input image $\bm{I}$ and the cell nodes embeddings $\bm{H}^{o}=[\bm{h}_1^{o}, \bm{h}_2^{o},\,\cdots \,,\bm{h}_{|\mathcal{V}|}^{o}]^\top$. Here, to emphasize the absolute position of cells, we calculate sinusoidal position embeddings~\cite{vaswani2017attention} using coordinates of cells and add them to $\bm{H}^{o}$. With $\bm{e}$ and $\bm{H}^{o}$ as the input sequence, the encoder Transformer calculates self-attention of the embedding sequences and generates encoding matrices. The encoding matrices are then used by the decoder to calculate cross-attention with word sequences. In the second stage, we use a decoder Transformer to autoregressively generate a report $R=[w_1, w_2,\,\cdots\,,w_z]$ for an input image within $z$ steps, where $w_i$ denotes a word. In step $t$, the decoder takes word sequence $R_{t-1}=[w_1, w_2,\,\cdots\,,w_{t-1}]$ as input to calculate self-attention and cross-attention for generating $w_t$. The decoder stops the generation process after generating an end token or $t>z$. 

The teacher forcing approach~\cite{williams1989learning} is applied in the training stage, where $R_{t-1}$ provided for the decoder is ground truth. Masked attention is applied to ensure the unprocessed positions to keep unseen from the model. In the testing stage, $R_{t-1}$ provided for the decoder is generated by the model itself in previous steps.

\begin{table*}
	\caption{Performance of text generation on NMI-WSI dataset ($\times$100). }\label{mainResult}
	\centering
	\begin{tabular}{c|c|c|c|c|c|c}
		\hline
		Model &BLEU-1 $\uparrow$ &  BLEU-4 $\uparrow$ & ROUGE $\uparrow$ & METEOR $\uparrow$ & CIDEr $\uparrow$ & SPICE $\uparrow$\\
		\hline
		MDNet&80.9$\pm$0.8&55.9$\pm$1.2&59.4$\pm$0.9&34.0$\pm$0.7&82.9$\pm$4.5&35.3$\pm$1.1\\
		UpDown-grid& 80.2$\pm$0.7 & 56.6$\pm$1.1& 60.4$\pm$0.7&34.0$\pm$0.3&103.6$\pm$2.5&36.5$\pm$0.1 \\ 
		SwinTrans&81.8$\pm$ 0.7&58.2$\pm$0.9&61.2$\pm$0.8&34.9$\pm$0.4&111.9$\pm$4.1&37.1$\pm$0.6\\
		GNNFormer & $\bm{83.0{\pm}0.9}$&$\bm{60.8{\pm}1.1}$&$\bm{63.0{\pm}0.9}$&$\bm{36.1{\pm}0.5}$&$\bm{126.9{\pm}3.1}$&$\bm{40.9{\pm}0.4}$\\
		\hline
	\end{tabular}
\end{table*}

\begin{table*}
	\caption{Performance of lesion recognition on NMI-WSI dataset ($\times$100)}\label{classifyResult}
	\centering
	\begin{tabular}{c|c|c|c|c}
		\hline
		Model & Accuracy $\uparrow$ & \thead{F1-Score\\macro} $\uparrow$ & \thead{F1-Score \\high grade} $\uparrow$ & 
		\thead{F1-Score \\low grade} $\uparrow$ \\
		\hline
        ResNet50 & 76.1$\pm$5.7 & 64.6$\pm$6.3 & 83.4$\pm$3.8 & 
        68.5$\pm$8.8 \\ 
	    Inception-V3 & 73.8$\pm$5.5 & 63.7$\pm$4.4 & 81.5$\pm$3.7 & 64.7$\pm$6.9 \\ 
	    DenseNet121 & 76.0$\pm$4.9 & 64.5$\pm$5.5 & 84.0$\pm$3.5 & 71.9$\pm$4.4 \\
	    MDNet &  70.4$\pm$5.6 & 63.0$\pm$5.0 & 78.0$\pm$4.2 & 64.6$\pm$5.4\\
	    UpDown-grid & 72.3$\pm$5.4 & 61.0$\pm$4.4 & 81.7$\pm$4.6 & 57.6$\pm$7.7\\
	    SwinTrans & 75.0$\pm$4.6 & 65.7$\pm$4.6 & 82.3$\pm$3.4 & 65.1$\pm$3.9 \\
	    GNNFormer & $\bm{79.1{\pm} 5.8}$ & $\bm{70.6{\pm}5.4}$ & $\bm{85.4{\pm}4.2}$ & $\bm{73.2{\pm}6.0}$ \\
		\hline
	\end{tabular}
\end{table*}

\section{Experiment}
We implement all methods with Pytorch~\footnote{https://pytorch.org/.}. We run all experiments on a workstation with an Intel (R) CPU E5-2620V4@2.1G of 16 cores, 128G RAM, and an NVIDIA (R) GPU TITAN XP with 12GB graphics memory.
\subsection{Dataset and Evaluation Metric}
We evaluate GNNFormer and baselines on the NMI-WSI dataset~\cite{zhang2019pathologist}~\footnote{Publicly available in https://github.com/zizhaozhang/nmi-wsi-diagnosis.}. \mbox{NMI-WSI} contains 4,253 tiles of size $1,024 \times 1,024$ cropped from 221 Whole-Slide Images~(WSIs) of non-invasive papillary urothelial carcinoma. Each tile is provided with 5 pieces of reports. 
Following the original partition~\cite{zhang2019pathologist}, we use 2,364 tiles from 113 WSIs for training and 1,889 tiles from 108 WSIs for validation and testing. Since no specific split is given for validation and testing, we randomly split the validation and testing dataset, with each dataset containing tiles from 54 WSIs. We repeat the split for 5 times, and report the mean of results.

 We adopt five widely-used evaluation metrics for image caption, including BLEU~\cite{papineni2002bleu}, ROUGE~\cite{lin2004rouge}, METEOR \cite{denkowski:lavie:meteor-wmt:2014}, CIDEr~\cite{vedantam2015cider} and SPICE~\cite{anderson2016spice}, to evaluate the performance of text generation of our method and the baselines. In particular, we adopt an open-source evaluation toolkit~\footnote{https://github.com/tylin/coco-caption.} to compute the metrics. 
 
Furthermore, we measure the performance of lesion recognition. In particular, we verify whether the generated reports give accurate diagnoses of the existence and degree of cancer areas.
Since this diagnosis is given as the final conclusion in each cytopathology report, we extract the conclusions of the ground-truth reports, and assign labels for corresponding image patches. 
In the NMI-WSI dataset, there are four types of conclusions, including low grade cancer, high grade cancer, normal and insufficient information. We merge image patches of ``normal'' and ``insufficient information'' into one class, resulting in three classes. There are two reasons for the merging. First, the two classes both do not possess obvious lesion areas. Image patches with no lesion detected are naturally categorized into the merged class. Second, the number of samples belonging to the two classes is small, making it hard to accurately distinguish between them. 
After the assignment of labels, we extract the conclusions of generated reports and compare them to the labels.
Accuracy and macro F1-Score are reported to show the overall performance on three classes. 
To show the diagnostic correctness on different lesion types, we also respectively treat images of high grade cancer and low grade cancer as positive samples, and report the F1-Scores.
 

\subsection{Baseline}

We compare GNNFormer with three report generation baseline models to demonstrate its effectiveness. 
The first baseline we choose is \mbox{MDNet}~\cite{conf/cvpr/ZhangXXMY17}, since it is a representative published work for cytopathology report generation with a reliable open-source implementation~\footnote{https://github.com/zizhaozhang/nmi-wsi-diagnosis}. 
The second baseline UpDown-grid is constructed by adapting the framework proposed in \cite{anderson2018bottom}. As in the original implementation, we apply ResNet101 to extract image feature maps, followed by a two-layer LSTM for top-down visual attention and caption generation, respectively. The third baseline SwinTrans uses the state-of-the-art Swin Transformer\cite{liu2021swin} for automatic generation of cytopathology reports. It adopts an encoder-decoder Transformer structure, with Swin-Transformer-Small as the encoder. The decoder has identical structure as that of GNNFormer. 

We also build image classification models as additional baselines to analyze the performance of lesion recognition. Three state-of-the-art models are adopted, including ResNet50, Inception-v3 and DenseNet121.

In GNNFormer, we use ResNet34~\cite{he2016deep} as feature extraction modules for cell nodes and the global images. We resize the pathology images to $500 \times 500$ before feeding them into the framework. We set the layer number of GNN to be $4$. For each GNN layer, we set the layer number of the MLP to be $2$. We set the layer number of both encoder and decoder Transformers to be $3$, head number to be 8, and the hidden dimension to be $256$. We use cross-entropy loss for optimization and optimize the model with Adam~\cite{DBLP:journals/corr/KingmaB14}. In the validation stage, we evaluate model performance according to the BLEU-4 score.

\begin{figure*}[t]
    \centering
	\includegraphics[width=0.95\linewidth]{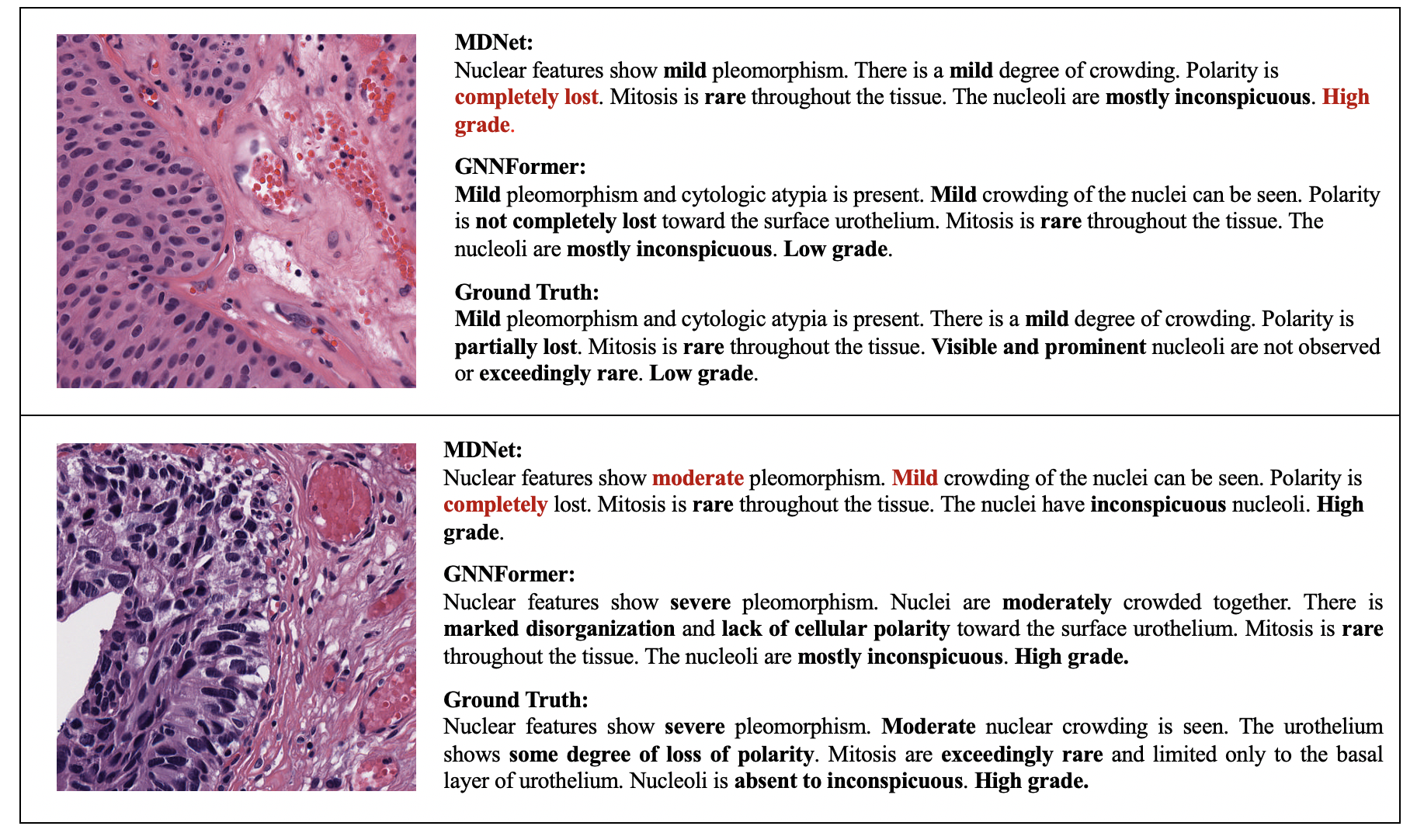}
	\caption{Example reports generated by MDNet and GNNFormer. The vital information is shown in boldface. And the inaccurate predictions are marked in red.} \label{generate}
	\vskip -0.1in
\end{figure*}
\begin{table*}[t] 
    \centering
	\caption{Ablation study ($\times$100).}\label{ablation1}
	\footnotesize
	\resizebox{\linewidth}{!}{\begin{tabular}{c|c|c|c|c|c|c|c|c}
	    \hline
		Method &BLEU-1 $\uparrow$ &  BLEU-4 $\uparrow$ & ROUGE $\uparrow$ & METEOR $\uparrow$ & CIDEr $\uparrow$ & SPICE $\uparrow$ & Accuracy $\uparrow$ & \thead{F1-Score\\macro} $\uparrow$\\
		\hline
		w/o Graph & 80.6$\pm$1.0&56.8$\pm$1.1&59.9$\pm$0.9&34.6$\pm$0.5&99.9$\pm$3.8&37.0$\pm$0.7 & 70.3$\pm$ 5.0 & 60.6$\pm$6.9\\
		w/o GNN &81.9$\pm$0.6&58.4$\pm$1.0&61.1$\pm$0.8&34.6$\pm$0.4&112.6$\pm$3.9&37.0$\pm$0.8 & 78.1$\pm$ 6.1 & 69.3$\pm$6.5\\
		Frozen CNN& 80.5$\pm$0.6&55.1$\pm$0.2&59.3$\pm$0.1&34.0$\pm$0.2&92.0$\pm$6.5&36.7$\pm$0.5 & 69.5$\pm$2.7 & 60.5$\pm$5.3\\
		w/o Global Embed & 80.4$\pm$0.8 & 57.9$\pm$1.0& 60.8$\pm$0.6&34.5$\pm$0.4&108.0$\pm$3.3&37.7$\pm$0.3&71.6$\pm$4.8&60.5$\pm$4.5 \\
		GNNFormer & $\bm{83.0{\pm}0.9}$&$\bm{60.8{\pm}1.1}$&$\bm{63.0{\pm}0.9}$&$\bm{36.1{\pm}0.5}$&$\bm{126.9{\pm}3.1}$&$\bm{40.9{\pm}0.4}$ & $\bm{79.1{\pm} 5.8}$ & $\bm{70.6{\pm}5.4}$\\
		\hline
	\end{tabular}}
	\vskip -0.1in
\end{table*}

\subsection{Results}
Results of text generation are summarized in Table~\ref{mainResult}. The best performance is shown in boldface. 
GNNFormer outperforms all baseline methods under the given evaluation metrics, showing GNNFormer's superiority in generating high-quality reports. In particular, the advantage of \mbox{GNNFormer} is more pronounced on the CIDEr and SPICE metrics, achieving 13.4\% and 10.2\% improvement compared with the second best results respectively. These two metrics are specially designed for image caption tasks. CIDEr puts larger weight on informative tokens that describe the degree of each reported phenomenon. SPICE uses ``scene-graph'' to measure how
effectively the generated reports recover objects, attributes, and relations described in ground truth. It is more tolerant of different expressions with the same meanings.

Table~\ref{classifyResult} shows the results of lesion recognition, which demonstrates the diagnostic correctness of each model. Though state-of-the-art classification models can achieve competitive performance compared to other report generation methods, GNNFormer achieves the best classification performance, with accuracy of 79.1\% and F1-Score of 70.6\%. GNNFormer also achieves the best results in diagnosing lesion types, with F1-Score for high grade cancer being 85.4\% and F1-Score for low grade cancer being 73.2\%.

Figure~\ref{generate} gives some example reports generated by \mbox{MDNet} and \mbox{GNNFormer}, together with the ground truth for comparison. Both methods can generate cytopathology reports with complete structure, while GNNFormer is more accurate in describing the degree of each phenomenon. 

\begin{figure*}[t]
    \centering
	\includegraphics[width=0.9\linewidth]{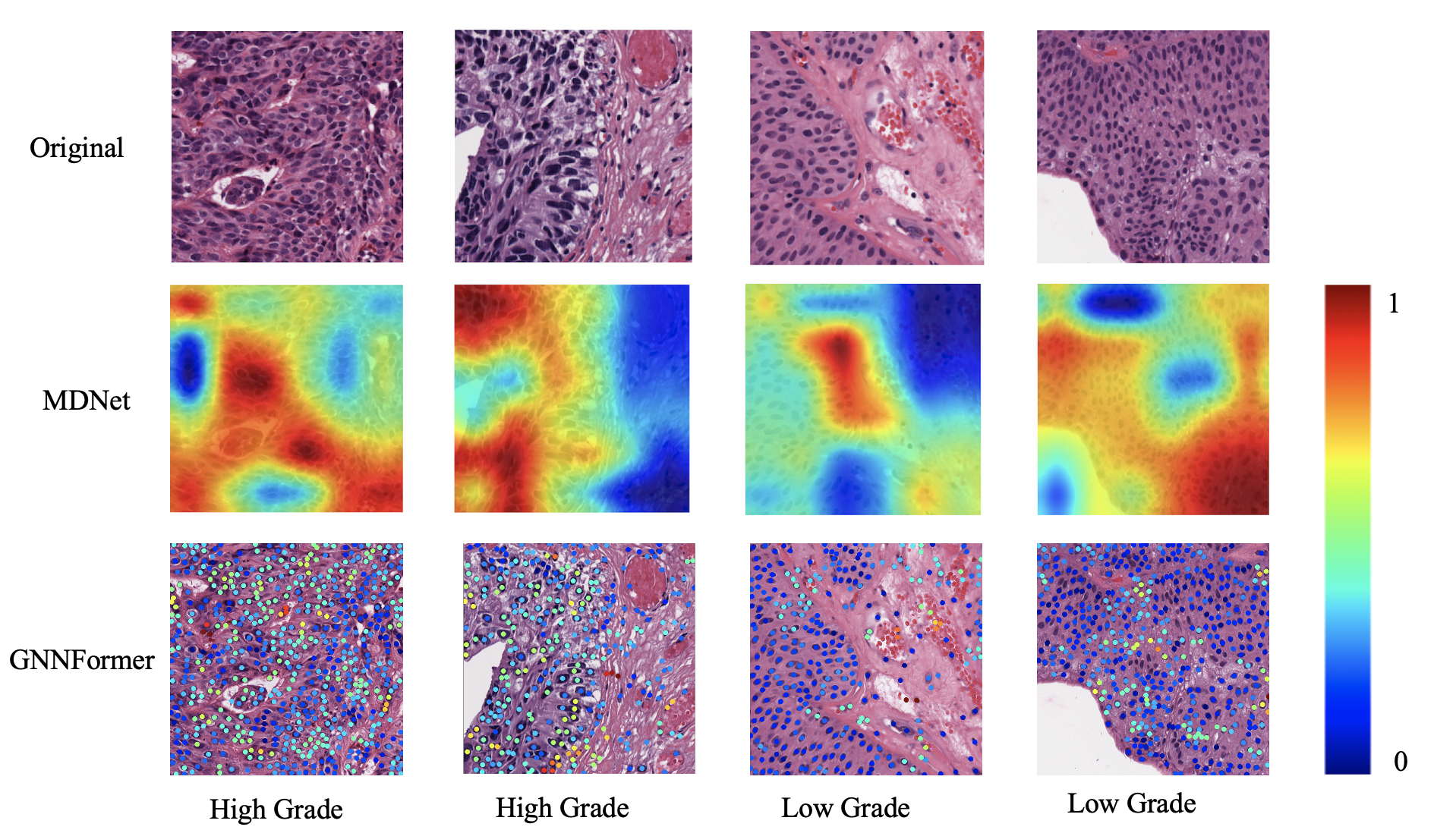}
	\caption{Interpretation with visualization. We show some visualization examples of MDNet and GNNFormer. Importance scores are standardized to be in [0,1] and are represented by different colors, which are denoted by the color bars on the right.} \label{viz}
	\vskip -0.1in
\end{figure*}

\subsection{Ablation Study}
We conduct ablation study to verify the effectiveness of different components in GNNFormer, including cell graph construction module, GNN module, global embedding extraction module and the trainable cell node morphology feature encoder. The results are summarized in Table~\ref{ablation1}. We describe the details of each model in the following.

``w/o Graph'' does not construct cell graphs and uses ResNet34 network to extract grid embeddings of input images. The grid embeddings are directly input into the Transformer encoder. ``w/o GNN'' constructs cell graphs and uses ResNet34 network to generate cell nodes embedding sequences $\bm{A}$ that captures morphology features. It directly inputs $\bm{A}$ into Transformer encoder, without applying GNN module to capture the structural information among cells.
``Frozen CNN'' freezes the parameters of the ResNet34 network without updating during the training process. ``w/o Global Embed'' constructs cell graphs and uses GNN for cell nodes embedding, but it does not calculate global embedding $\bm{e}$ of input images. 

We can draw the following conclusions: 
\begin{itemize}
    \item By comparing ``w/o GNN'' with ``w/o Graph'', we can find that generating cell node embeddings instead of grid embeddings can improve the performance of report generation.
    \item By comparing ``w/o GNN'' with GNNFormer, we can find that structural information among cells is important for generating high-quality reports. 
    \item By comparing ``w/o Global Embed" with GNNFormer, we can find that global embedding of the image is also important for achieving good performance. 
    \item By comparing ``Frozen CNN" with GNNFormer, we can find that an adaptive CNN feature extractor for cell nodes with parameter updating can boost the model performance. 
\end{itemize}

\subsection{Interpretation with Visualization}
With the method proposed in~\cite{Chefer_2021_ICCV}, the importance score of each cell node for generating the final report can be calculated. According to the scores, we generate visualization for cell nodes to provide intuitive interpretation of the report-generating process for pathologists. Although \mbox{MDNet} also provides Region-Of-Interest visualization based on attention score, visualization of GNNFormer is more detailed due to its cell-based scoring ability. Figure~\ref{viz} gives some examples. Note that visualization for every generated word is obtainable. Due to space limit, we only show the visualization for the diagnosis conclusion~(``high grade cancer" or ``low grade cancer"), which is the most important part. We can observe that in images with high grade cancer, cells with irregular shapes and chaotically arranged are given higher scores, which means they are more likely to be noticed by the model when making conclusion. In images with low grade cancer, cells with more rounded shapes while still disorderly arranged are given higher scores. This is consistent with the actual judgement criteria, making the visualization understandable for pathologists.

\subsection{Influence of Morphology Feature Encoder}
\begin{figure}[t]
    \centering
	\includegraphics[width=0.8\linewidth]{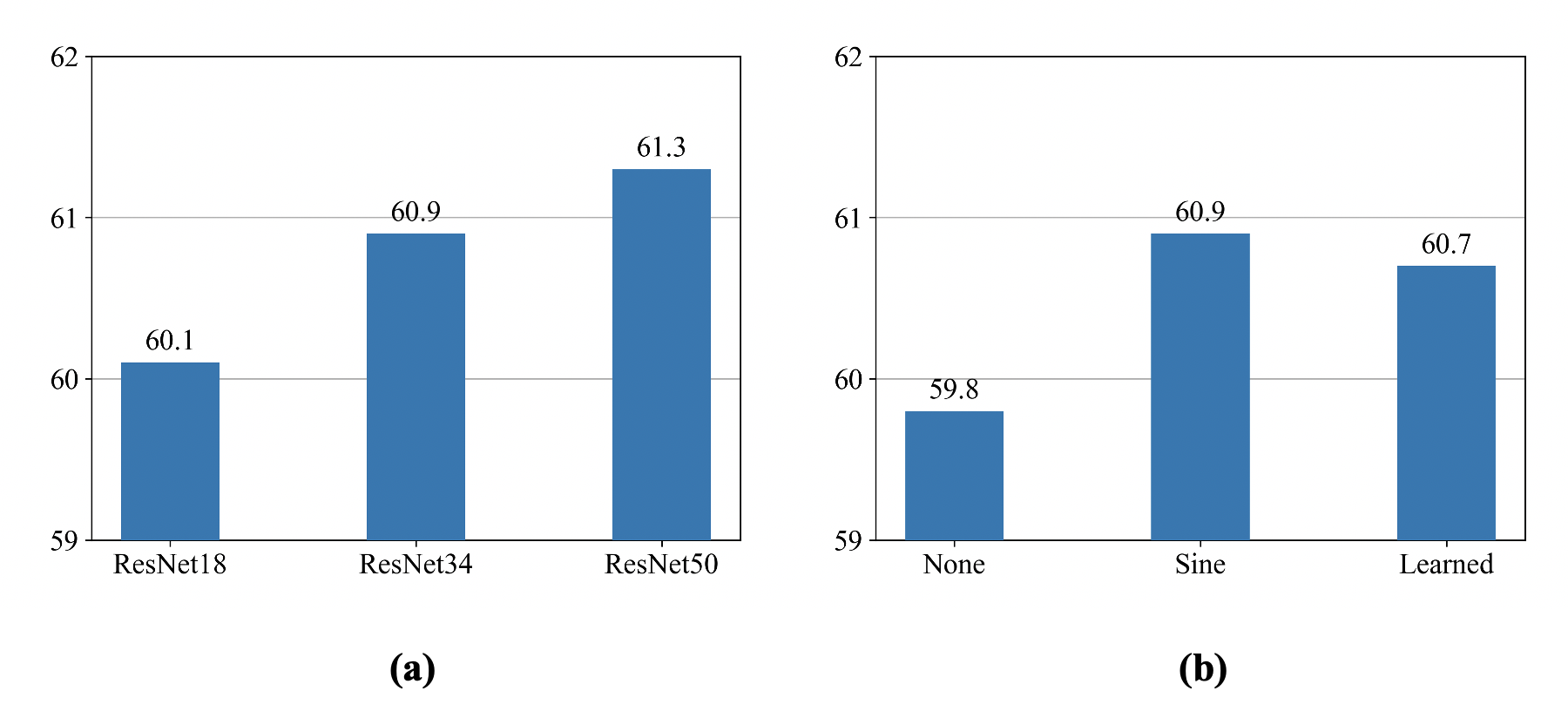}
	\caption{(a) Influence of morphology feature encoder. We show BLEU-4 scores on validation set for different morphology feature encoders. (b) Influence of position embedding. We show BLEU-4 scores on validation set for all hyper-parameters.
 } \label{influence}
    \vskip -0.1in
\end{figure}

In this subsection, we discuss the choice of the morphology feature encoder for cell nodes, and its influence on \mbox{GNNFormer}'s performance. We respectively apply ResNet18, ResNet34 and ResNet50 as the morphology feature encoder for cell nodes. The BLEU-4 scores on validation set are reported in Figure~\ref{influence} (a). We can find that as the layer number of ResNet grows, GNNFormer tends to perform better. We did not try all state-of-the-art image feature encoding models since this is not the focus of \mbox{GNNFormer}. To keep the model complexity moderate, we choose ResNet34 as the morphology feature encoder in \mbox{GNNFormer}. The performance of GNNFormer could be further improved with other image feature encoding models, which is worth exploring in future study.

\subsection{Influence of Position Embedding}
Adding position embedding for cell nodes is a technique we use to emphasize the global coordinates of cell nodes. We have adapted two different kinds of position embedding methods, including ``Sine" and ``Learned", for \mbox{GNNFormer}. ``Sine" method uses sine and cosine functions to calculate embedding vectors given row or column coordinates. ``Learned" method learns embedding matrices for row and column coordinates, respectively. We also give the result of a variant which does not use position embedding. This variant is called ``None". The BLEU-4 scores on the validation set are shown in Figure~\ref{influence} (b). The degraded performance of the variant with ``None" position embedding shows the effectiveness of position embedding. We also find that the performance of GNNFormer with ``Learned" method is not significantly better than that with ``Sine" method, but brings additional parameters. Hence, ``Sine" method is adopted in GNNFormer. Better performance could be expected with more advanced position embedding methods. But this is not the focus of this paper and is left for future study. 

\subsection{Sensitivity to Hyper-Parameters}
\begin{figure}[t]
    \centering	
    \includegraphics[width=0.8\linewidth]{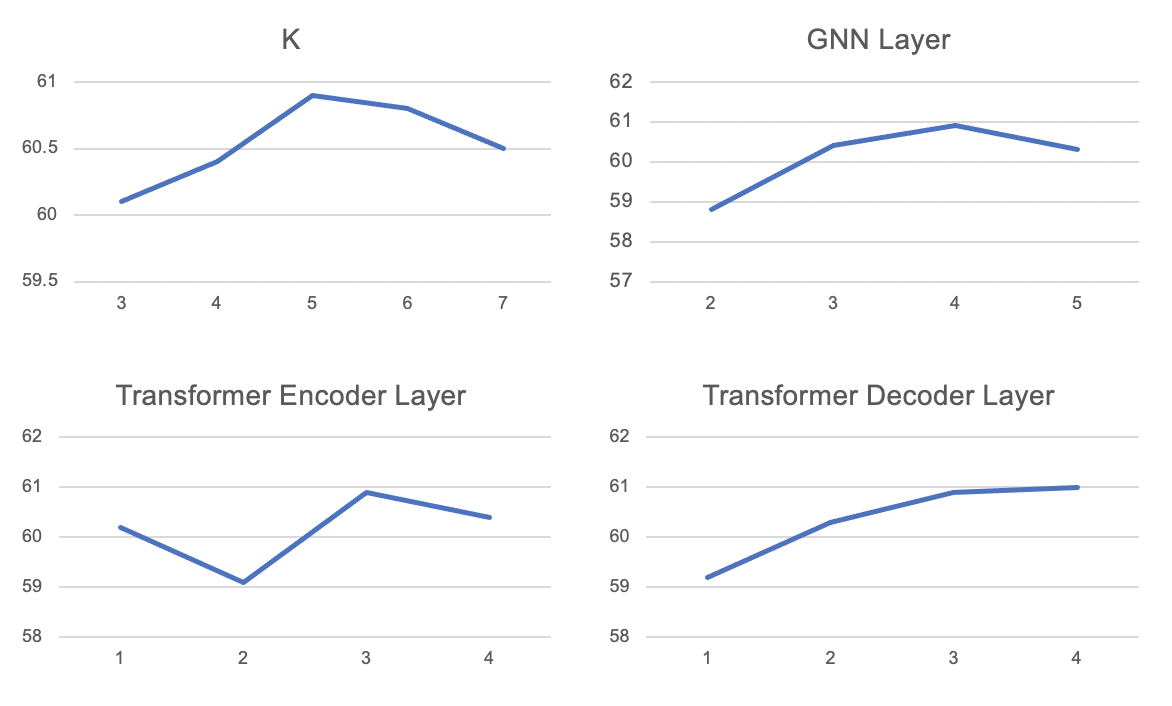}
	\caption{Sensitivity to hyper-parameters. We show BLEU-4 scores on validation set for all hyper-parameters.} \label{hyper}
	\vskip -0.05in
\end{figure}

In this subsection, we analyze the sensitivity to hyper-parameters, including $k$ in the kNN algorithm, layer number of GNN module, layer number of Transformer encoder and layer number of Transformer decoder. Results on validation set are shown in Figure~\ref{hyper}.

For $k$ in kNN algorithm, we can observe that \mbox{GNNFormer} behaves well in the range of $[4, 7]$. With respect to the layer number of GNN module, we can find that \mbox{GNNFormer} achieves good performance within the range of $\{3,4,5\}$. As for the layer number of Transformer encoder, we can see that \mbox{GNNFormer} performs steadily with $\{3,4,5\}$. As for the layer number of Transformer decoder, we can also observe that \mbox{GNNFormer} performs steadily with $\{3,4\}$.

\section{Conclusion}
In this paper, we propose a novel graph-based framework called \mbox{GNNFormer}, which seamlessly integrates GNN and Transformer into the same framework, for cytopathology report generation. GNNFormer explicitly models the structural information among cells by constructing a cell graph. It also effectively fuses the structural information with fine-grained morphology features and background features to generate high-quality reports. Experiments on the NMI-WSI dataset show that GNNFormer can outperform other state-of-the-art baselines on cytopathology report generation. Furthermore, GNNFormer naturally generates a visualization of cell's importance scores, which provides intuitive interpretation of the report-generating process for pathologists. 

\section*{Limitations} 
We only use the NMI-WSI dataset for evaluation, because to the best of our knowledge NMI-WSI is the only dataset which is publicly available with standard cytopathology reports. Fortunately, NMI-WSI  is from a solid and influential research, and has high quality. 

{\small
\bibliographystyle{unsrt}
\bibliography{ref}

\begin{thebibliography}{10}

\bibitem{graham2019hover}
Simon Graham, Quoc~Dang Vu, Shan E~Ahmed Raza, Ayesha Azam, Yee~Wah Tsang,
  Jin~Tae Kwak, and Nasir Rajpoot.
\newblock Hover-net: Simultaneous segmentation and classification of nuclei in
  multi-tissue histology images.
\newblock {\em Medical Image Analysis}, 58:101563, 2019.

\bibitem{wang2020single}
Haiyue Wang, Yuming Jiang, Bailiang Li, Yi~Cui, Dengwang Li, and Ruijiang Li.
\newblock Single-cell spatial analysis of tumor and immune microenvironment on
  whole-slide image reveals hepatocellular carcinoma subtypes.
\newblock {\em Cancers}, 12(12):3562, 2020.

\bibitem{liang2018weakly}
Qiaokang Liang, Yang Nan, Gianmarc Coppola, Kunglin Zou, Wei Sun, Dan Zhang,
  Yaonan Wang, and Guanzhen Yu.
\newblock Weakly supervised biomedical image segmentation by reiterative
  learning.
\newblock {\em IEEE Journal of Biomedical and Health Informatics},
  23(3):1205--1214, 2018.

\bibitem{lu2021ai}
Ming~Y Lu, Tiffany~Y Chen, Drew~FK Williamson, Melissa Zhao, Maha Shady, Jana
  Lipkova, and Faisal Mahmood.
\newblock Ai-based pathology predicts origins for cancers of unknown primary.
\newblock {\em Nature}, 594(7861):106--110, 2021.

\bibitem{conf/cvpr/ZhangXXMY17}
Zizhao Zhang, Yuanpu Xie, Fuyong Xing, Mason McGough, and Lin Yang.
\newblock Mdnet: {A} semantically and visually interpretable medical image
  diagnosis network.
\newblock In {\em IEEE Conference on Computer Vision and Pattern Recognition},
  pages 3549--3557, 2017.

\bibitem{he2016deep}
Kaiming He, Xiangyu Zhang, Shaoqing Ren, and Jian Sun.
\newblock Deep residual learning for image recognition.
\newblock In {\em IEEE Conference on Computer Vision and Pattern Recognition},
  pages 770--778, 2016.

\bibitem{jing2018automatic}
Baoyu Jing, Pengtao Xie, and Eric Xing.
\newblock On the automatic generation of medical imaging reports.
\newblock In {\em Annual Meeting of the Association for Computational
  Linguistics}, pages 2577--2586, 2018.

\bibitem{zhang2019pathologist}
Zizhao Zhang, Pingjun Chen, Mason McGough, Fuyong Xing, Chunbao Wang, Marilyn
  Bui, Yuanpu Xie, Manish Sapkota, Lei Cui, Jasreman Dhillon, et~al.
\newblock Pathologist-level interpretable whole-slide cancer diagnosis with
  deep learning.
\newblock {\em Nature Machine Intelligence}, 1(5):236--245, 2019.

\bibitem{lu2018feature}
Cheng Lu, Xiangxue Wang, Prateek Prasanna, German Corredor, Geoffrey Sedor,
  Kaustav Bera, Vamsidhar Velcheti, and Anant Madabhushi.
\newblock Feature driven local cell graph (fedeg): predicting overall survival
  in early stage lung cancer.
\newblock In {\em International Conference on Medical Image Computing and
  Computer-Assisted Intervention}, pages 407--416, 2018.

\bibitem{scarselli2008graph}
Franco Scarselli, Marco Gori, Ah~Chung Tsoi, Markus Hagenbuchner, and Gabriele
  Monfardini.
\newblock The graph neural network model.
\newblock {\em IEEE transactions on Neural Networks}, 20(1):61--80, 2008.

\bibitem{vaswani2017attention}
Ashish Vaswani, Noam Shazeer, Niki Parmar, Jakob Uszkoreit, Llion Jones,
  Aidan~N Gomez, {\L}ukasz Kaiser, and Illia Polosukhin.
\newblock Attention is all you need.
\newblock In {\em Advances in Neural Information Processing Systems}, pages
  1735--1780, 2017.

\bibitem{conf/cvpr/VinyalsTBE15}
Oriol Vinyals, Alexander Toshev, Samy Bengio, and Dumitru Erhan.
\newblock Show and tell: A neural image caption generator.
\newblock In {\em IEEE Conference on Computer Vision and Pattern Recognition},
  pages 3156--3164, 2015.

\bibitem{Donahue_2015_CVPR}
Jeffrey Donahue, Lisa Anne~Hendricks, Sergio Guadarrama, Marcus Rohrbach,
  Subhashini Venugopalan, Kate Saenko, and Trevor Darrell.
\newblock Long-term recurrent convolutional networks for visual recognition and
  description.
\newblock In {\em IEEE Conference on Computer Vision and Pattern Recognition},
  pages 2625--2634, 2015.

\bibitem{xu2015show}
Kelvin Xu, Jimmy Ba, Ryan Kiros, Kyunghyun Cho, Aaron Courville, Ruslan
  Salakhudinov, Rich Zemel, and Yoshua Bengio.
\newblock Show, attend and tell: Neural image caption generation with visual
  attention.
\newblock In {\em International Conference on Machine Learning}, pages
  2048--2057, 2015.

\bibitem{anderson2018bottom}
Peter Anderson, Xiaodong He, Chris Buehler, Damien Teney, Mark Johnson, Stephen
  Gould, and Lei Zhang.
\newblock Bottom-up and top-down attention for image captioning and visual
  question answering.
\newblock In {\em IEEE Conference on Computer Vision and Pattern Recognition},
  pages 6077--6086, 2018.

\bibitem{conf/eccv/YaoPLM18}
Ting Yao, Yingwei Pan, Yehao Li, and Tao Mei.
\newblock Exploring visual relationship for image captioning.
\newblock In {\em European Conference on Computer Vision}, pages 711--727,
  2018.

\bibitem{conf/cvpr/YangTZC19}
Xu~Yang, Kaihua Tang, Hanwang Zhang, and Jianfei Cai.
\newblock Auto-encoding scene graphs for image captioning.
\newblock In {\em IEEE Conference on Computer Vision and Pattern Recognition},
  pages 10685--10694, 2019.

\bibitem{conf/acl/SoricutDSG18}
Piyush Sharma, Nan Ding, Sebastian Goodman, and Radu Soricut.
\newblock Conceptual captions: {A} cleaned, hypernymed, image alt-text dataset
  for automatic image captioning.
\newblock In {\em Annual Meeting of the Association for Computational
  Linguistics}, pages 2556--2565, 2018.

\bibitem{conf/nips/HerdadeKBS19}
Simao Herdade, Armin Kappeler, Kofi Boakye, and Joao Soares.
\newblock Image captioning: Transforming objects into words.
\newblock In {\em Advances in Neural Information Processing Systems}, pages
  11135--11145, 2019.

\bibitem{cornia2020meshed}
Marcella Cornia, Matteo Stefanini, Lorenzo Baraldi, and Rita Cucchiara.
\newblock Meshed-memory transformer for image captioning.
\newblock In {\em IEEE Conference on Computer Vision and Pattern Recognition},
  pages 10578--10587, 2020.

\bibitem{wang2016deep}
Dayong Wang, Aditya Khosla, Rishab Gargeya, Humayun Irshad, and Andrew~H Beck.
\newblock Deep learning for identifying metastatic breast cancer.
\newblock {\em CoRR}, abs/1606.05718, 2016.

\bibitem{garcia2017automatic}
Emilio Garcia, Renato Hermoza, Cesar~Beltran Castanon, Luis Cano, Miluska
  Castillo, and Carlos Castanneda.
\newblock Automatic lymphocyte detection on gastric cancer ihc images using
  deep learning.
\newblock In {\em IEEE International Symposium on Computer-based Medical
  Systems}, pages 200--204, 2017.

\bibitem{li2018deep}
Yuexiang Li, Xuechen Li, Xinpeng Xie, and Linlin Shen.
\newblock Deep learning based gastric cancer identification.
\newblock In {\em IEEE International Symposium on Biomedical Imaging}, pages
  182--185, 2018.

\bibitem{raju2020graph}
Ashwin Raju, Jiawen Yao, Mohammad~MinHazul Haq, Jitendra Jonnagaddala, and
  Junzhou Huang.
\newblock Graph attention multi-instance learning for accurate colorectal
  cancer staging.
\newblock In {\em International Conference on Medical Image Computing and
  Computer-Assisted Intervention}, pages 529--539, 2020.

\bibitem{anand2020histographs}
Deepak Anand, Shrey Gadiya, and Amit Sethi.
\newblock Histographs: graphs in histopathology.
\newblock In {\em Medical Imaging: Digital Pathology}, volume 11320, pages
  150--155, 2020.

\bibitem{graham2019mild}
Simon Graham, Hao Chen, Jevgenij Gamper, Qi~Dou, Pheng-Ann Heng, David Snead,
  Yee~Wah Tsang, and Nasir Rajpoot.
\newblock Mild-net: Minimal information loss dilated network for gland instance
  segmentation in colon histology images.
\newblock {\em Medical Image Analysis}, 52:199--211, 2019.

\bibitem{zhang2022deep}
Song Zhang, Yangfan Zhou, Dehua Tang, Muhan Ni, Jinyu Zheng, Guifang Xu,
  Chunyan Peng, Shanshan Shen, Qiang Zhan, Xiaoyun Wang, et~al.
\newblock A deep learning-based segmentation system for rapid onsite cytologic
  pathology evaluation of pancreatic masses: A retrospective, multicenter,
  diagnostic study.
\newblock {\em EBioMedicine}, 80:104022, 2022.

\bibitem{zhu2017wsisa}
Xinliang Zhu, Jiawen Yao, Feiyun Zhu, and Junzhou Huang.
\newblock Wsisa: Making survival prediction from whole slide histopathological
  images.
\newblock In {\em IEEE Conference on Computer Vision and Pattern Recognition},
  pages 7234--7242, 2017.

\bibitem{mobadersany2018predicting}
Pooya Mobadersany, Safoora Yousefi, Mohamed Amgad, David~A Gutman, Jill~S
  Barnholtz-Sloan, Jos{\'e}~E Vel{\'a}zquez~Vega, Daniel~J Brat, and Lee~AD
  Cooper.
\newblock Predicting cancer outcomes from histology and genomics using
  convolutional networks.
\newblock {\em National Academy of Sciences}, 115(13):E2970--E2979, 2018.

\bibitem{lu2021data}
Ming~Y Lu, Drew~FK Williamson, Tiffany~Y Chen, Richard~J Chen, Matteo Barbieri,
  and Faisal Mahmood.
\newblock Data-efficient and weakly supervised computational pathology on
  whole-slide images.
\newblock {\em Nature Biomedical Engineering}, 5(6):555--570, 2021.

\bibitem{li2018graph}
Ruoyu Li, Jiawen Yao, Xinliang Zhu, Yeqing Li, and Junzhou Huang.
\newblock Graph cnn for survival analysis on whole slide pathological images.
\newblock In {\em International Conference on Medical Image Computing and
  Computer-Assisted Intervention}, pages 174--182, 2018.

\bibitem{chen2021whole}
Richard~J Chen, Ming~Y Lu, Muhammad Shaban, Chengkuan Chen, Tiffany~Y Chen,
  Drew~FK Williamson, and Faisal Mahmood.
\newblock Whole slide images are 2d point clouds: Context-aware survival
  prediction using patch-based graph convolutional networks.
\newblock In {\em International Conference on Medical Image Computing and
  Computer-Assisted Intervention}, pages 339--349, 2021.

\bibitem{jaume2021quantifying}
Guillaume Jaume, Pushpak Pati, Behzad Bozorgtabar, Antonio Foncubierta,
  Anna~Maria Anniciello, Florinda Feroce, Tilman Rau, Jean-Philippe Thiran,
  Maria Gabrani, and Orcun Goksel.
\newblock Quantifying explainers of graph neural networks in computational
  pathology.
\newblock In {\em IEEE Conference on Computer Vision and Pattern Recognition},
  volume~12, page 3562, 2021.

\bibitem{gamper2019pannuke}
Jevgenij Gamper, Navid Alemi~Koohbanani, Ksenija Benet, Ali Khuram, and Nasir
  Rajpoot.
\newblock Pannuke: an open pan-cancer histology dataset for nuclei instance
  segmentation and classification.
\newblock In {\em European Congress on Digital Pathology}, volume 11435, pages
  11--19, 2019.

\bibitem{deng2009imagenet}
Jia Deng, Wei Dong, Richard Socher, Li-Jia Li, Kai Li, and Li~Fei-Fei.
\newblock Imagenet: a large-scale hierarchical image database.
\newblock In {\em IEEE Conference on Computer Vision and Pattern Recognition},
  pages 248--255, 2009.

\bibitem{DBLP:conf/iclr/XuHLJ19}
Keyulu Xu, Weihua Hu, Jure Leskovec, and Stefanie Jegelka.
\newblock How powerful are graph neural networks?
\newblock In {\em International Conference on Learning Representations}, 2019.

\bibitem{williams1989learning}
Ronald~J Williams and David Zipser.
\newblock A learning algorithm for continually running fully recurrent neural
  networks.
\newblock {\em Neural Computation}, 1(2):270--280, 1989.

\bibitem{papineni2002bleu}
Kishore Papineni, Salim Roukos, Todd Ward, and Wei-Jing Zhu.
\newblock Bleu: a method for automatic evaluation of machine translation.
\newblock In {\em Annual Meeting of the Association for Computational
  Linguistics}, pages 1106--1114, 2002.

\bibitem{lin2004rouge}
Chin-Yew Lin.
\newblock Rouge: a package for automatic evaluation of summaries.
\newblock In {\em Text Summarization Branches Out}, pages 74--81, 2004.

\bibitem{denkowski:lavie:meteor-wmt:2014}
Michael Denkowski and Alon Lavie.
\newblock Meteor universal: language specific translation evaluation for any
  target language.
\newblock In {\em EACL Workshop on Statistical Machine Translation}, pages
  376--380, 2014.

\bibitem{vedantam2015cider}
Ramakrishna Vedantam, C~Lawrence~Zitnick, and Devi Parikh.
\newblock Cider: consensus-based image description evaluation.
\newblock In {\em IEEE Conference on Computer Vision and Pattern Recognition},
  pages 4566--4575, 2015.

\bibitem{anderson2016spice}
Peter Anderson, Basura Fernando, Mark Johnson, and Stephen Gould.
\newblock Spice: semantic propositional image caption evaluation.
\newblock In {\em European Conference on Computer Vision}, pages 382--398,
  2016.

\bibitem{liu2021swin}
Ze~Liu, Yutong Lin, Yue Cao, Han Hu, Yixuan Wei, Zheng Zhang, Stephen Lin, and
  Baining Guo.
\newblock Swin transformer: hierarchical vision transformer using shifted
  windows.
\newblock In {\em International Conference on Computer Vision}, pages
  1724--1734, 2021.

\bibitem{DBLP:journals/corr/KingmaB14}
Diederik~P. Kingma and Jimmy Ba.
\newblock {A} method for stochastic optimization.
\newblock In {\em International Conference on Learning Representations}, 2015.

\bibitem{Chefer_2021_ICCV}
Hila Chefer, Shir Gur, and Lior Wolf.
\newblock Generic attention-model explainability for interpreting bi-modal and
  encoder-decoder transformers.
\newblock In {\em International Conference on Computer Vision}, pages 397--406,
  2021.

\end{thebibliography}
}

\end{document}